\documentclass{article}

\usepackage[final]{nips_2017}

\usepackage[utf8]{inputenc} 
\usepackage[T1]{fontenc}    
\usepackage{hyperref}       
\usepackage{url}            
\usepackage{booktabs}       
\usepackage{amsfonts}       
\usepackage{nicefrac}       
\usepackage{microtype}      
\usepackage{amsmath}
\usepackage[pdftex]{graphicx}
\usepackage{rotating}
\usepackage{multirow}

\title{SANTIS: Sampling-Augmented Neural neTwork with Incoherent Structure for MR image reconstruction}

\author{Fang Liu \\
Department of Radiology\\
University of Wisconsin-Madison\\
Madison, Wisconsin USA\\
  \texttt{leoliuf@gmail.com} \\
  \And
Lihua Chen\\
Department of Radiology\\
Southwest Hospital\\
Chongqing, China\\
  \And
Richard Kijowski\\
Department of Radiology\\
University of Wisconsin-Madison\\
Madison, Wisconsin USA\\
  \And
Li Feng\\
Department of Medical Physics\\
Memorial Sloan Kettering Cancer Center \\
New York, NY USA\\
}

\begin{document}

\maketitle

\begin{abstract}
\textbf{Purpose:} To develop and evaluate a novel deep learning-based reconstruction framework called SANTIS (Sampling-Augmented Neural neTwork with Incoherent Structure) for efficient MR image reconstruction with improved robustness against sampling pattern discrepancy.
\textbf{Methods: }SANTIS uses a data cycle-consistent adversarial network combining efficient end-to-end convolutional neural network mapping, data fidelity enforcement and adversarial training for reconstructing accelerated MR images more faithfully. A training strategy employing sampling augmentation with extensive variation of undersampling patterns was further introduced to promote the robustness of the trained network, which is capable of exploring a variety of aliasing artifact structures and thus can be applied to remove undersampling artifacts more faithfully. The performance of SANTIS was demonstrated for accelerated knee images using a Cartesian trajectory and for accelerated liver images using a golden-angle radial trajectory. Several quantitative metrics were used to assess its performance against fully sampled artifact-free images and different reference reconstruction approaches. The performance of SANTIS in reconstructing liver images at different contrast phases was also tested using transfer learning.
\textbf{Results: }Compared to conventional reconstruction that exploited image sparsity and standard deep learning methods without sampling augmentation (e.g., training with a fixed undersampling pattern), SANTIS achieved consistent better reconstruction performance for both knee and liver images, with lower errors, greater image sharpness and higher similarity with respect to the reference regardless of the undersampling patterns during inference. SANTIS also achieved encouraging results in reconstructing liver images acquired at different contrast phases.
\textbf{Conclusion: }By extensively varying undersampling patterns, the sampling-augmented training strategy in SANTIS can recognize various aliasing artifact structures and can be used to remove undersampling artifacts robustly. This novel concept behind SANTIS can particularly be useful towards improving the robustness of deep learning-based image reconstruction against discrepancy between training and evaluation, which is currently an important but less studied open question.

\end{abstract}

\section{INTRODUCTION}
In recent years, there has been an explosive growth of interest in applying deep learning to a wide range of medical imaging applications, including disease classification (1–3), tissue segmentation (4–8), and lesion detection and recognition (9–12), etc. In addition, deep learning also holds great promise in the reconstruction of undersampled MRI data, providing new opportunities to escalate the performance of rapid MRI. Many pioneer works, presented over the past few years (13–21), have all demonstrated that deep learning-based reconstruction methods could yield improved image quality with dramatically reduced reconstruction time. Compared to conventional techniques such as parallel imaging (22–24) and compressed sensing (25–27), deep learning-based reconstruction does not require a specific model assumption, eliminates the needs for a computationally-expensive iterative process, and aims to remove undersampling artifacts by inferencing features from a large image dataset. For example, a straightforward end-to-end mapping using convolutional neural networks (CNNs) has been demonstrated for efficiently removing artifacts from undersampled images (16–18), directly translating \textit{k}-space information into image (20), and faithfully estimating missing \textit{k}-space data (19,21). More importantly, upon completion of the one-time CNN training for a particular acquisition protocol, a deep learning system can reconstruct new undersampled images at an extremely fast speed, generally in order of seconds (13–21).

In existing deep learning-based reconstruction methods, supervised training is performed using artifact-free reference images and their corresponding undersampled pairs (13–21). The undersampled images are typically generated by a fixed undersampling pattern in the network training, and the trained network is then applied to reconstruct new images acquired with the same undersampling pattern in the inference step. While such a training strategy can maintain a favorable reconstruction performance for a pre-selected undersampling pattern, the robustness of the trained network against any discrepancy of undersampling schemes is typically poor. A recent study has shown that a deviation of undersampling patterns between the training and inference can degrade image quality, and the severity of the influence is dependent on a specific network structure and applied image datasets (28). Such an observation elaborates the importance of a better network training strategy, with which a system can be robust in removing image artifacts from a broad spectrum of undersampling patterns. 

The main contribution of this work was to propose a new training strategy to improve the robustness of a trained network against discrepancy of undersampling schemes that may occur between training and inference. The hypothesis was that deep learning-based reconstruction could benefit from extensively varying undersampling patterns during the training process, leading to better generalization and performance of the trained network in removing undersampling artifacts. Such a training strategy enforces sampling or \textit{k}-space trajectory augmentation, representing a great candidate for highly efficient and robust image reconstruction, and may be well-suited for applications that employ non-repeating \textit{k}-space undersampling. This new framework, termed as \underline{S}ampling-\underline{A}ugmented \underline{N}eural ne\underline{T}work with \underline{I}ncoherent \underline{S}tructure (SANTIS), was incorporated into a data cycle-consistent adversarial network and was evaluated in a retrospectively undersampled knee and liver MR datasets using variable-density Cartesian sampling and golden-angle radial sampling, respectively.

\section{THEORY}
\subsection{CNN-Regularized Image Reconstruction }

In MR image reconstruction, an image is estimated from a discrete set of \textit{k}-space measurements that can be expressed as:

\begin{equation} \label{eq1}
d = Ex + \varepsilon
\end{equation}

Here, $d$ is a vector of length $N$ for measured \textit{k}-space data and ${N = {N_x} \times {N_y}}$  is the total number of \textit{k}-space samples. $x$ is the image to be reconstructed, and $\varepsilon$ represents complex noise in the measurements. In a general case, the encoding matrix $E$ can be expanded as: 

\begin{equation} \label{eq2}
E = U\Im C
\end{equation}

where $\Im$ is the Fourier transform operator and $C$ represents coil sensitivities. In the case of undersampling, $U$ is an undersampling mask to select desired \textit{k}-space data points. An SNR-optimized reconstruction can be accomplished in the least squares sense as: 

\begin{equation} \label{eq3}
\hat x = \arg \;\mathop {\min }\limits_x \frac{1}{2}\left\| {Ex - d} \right\|_2^2
\end{equation}

where  ${\left\| . \right\|_2}$ is the $l_2$  norm. Eq.\ref{eq3} can be poorly-conditioned with a high undersampling factor, and thus, a regularization term can be additionally incorporated as: 

\begin{equation} \label{eq4}
\hat x = \arg \;\mathop {\min }\limits_x \left( {\frac{1}{2}\left\| {Ex - d} \right\|_2^2 + \lambda \Re \left( x \right)} \right)
\end{equation}

The design of the penalty function $\Re$ is governed by prior knowledge about the image $x$ , with a regularization parameter $\lambda$  controlling the trade-off between the data fidelity (the left term) and prior information assumptions (the right term). In conventional MR reconstruction, $\Re$  is often chosen as 

\begin{equation} \label{eq5}
\Re \left( x \right) = {\left\| {{\rm T}x} \right\|_N}
\end{equation}

where ${\left\| . \right\|_N}$  represents an N-norm and T is a specific transform operator, which, for example, is often selected as the wavelet transform or finite differences in the context of compressed sensing (25). 

In contrast to conventional regularizations, the penalty function can be implemented by learning an end-to-end deep CNN mapping from an undersampled image to its corresponding fully sampled pair, like many recent proposals (13–15,17). In other words, a CNN can serve as a spatial or spatial-temporal regularization term in constrained image reconstruction, and Eq.\ref{eq4} can be rewritten as: 

\begin{equation} \label{eq6}
\hat x = \arg \;\mathop {\min }\limits_x \left( {\frac{1}{2}\left\| {Ex - d} \right\|_2^2 + \lambda {{\left\| {F({x_u}) - x} \right\|}_N}} \right)
\end{equation}

Here, ${F:{f_{gen}}\left( {{x_u}|\theta } \right) \to \hat x}$ is a generator function using a deep CNN conditioned on network parameters $\theta$, and the ${x_u}$ is an undersampled image to be reconstructed. 

\begin{figure}[h]
  \centering
  \includegraphics[width=1\linewidth]{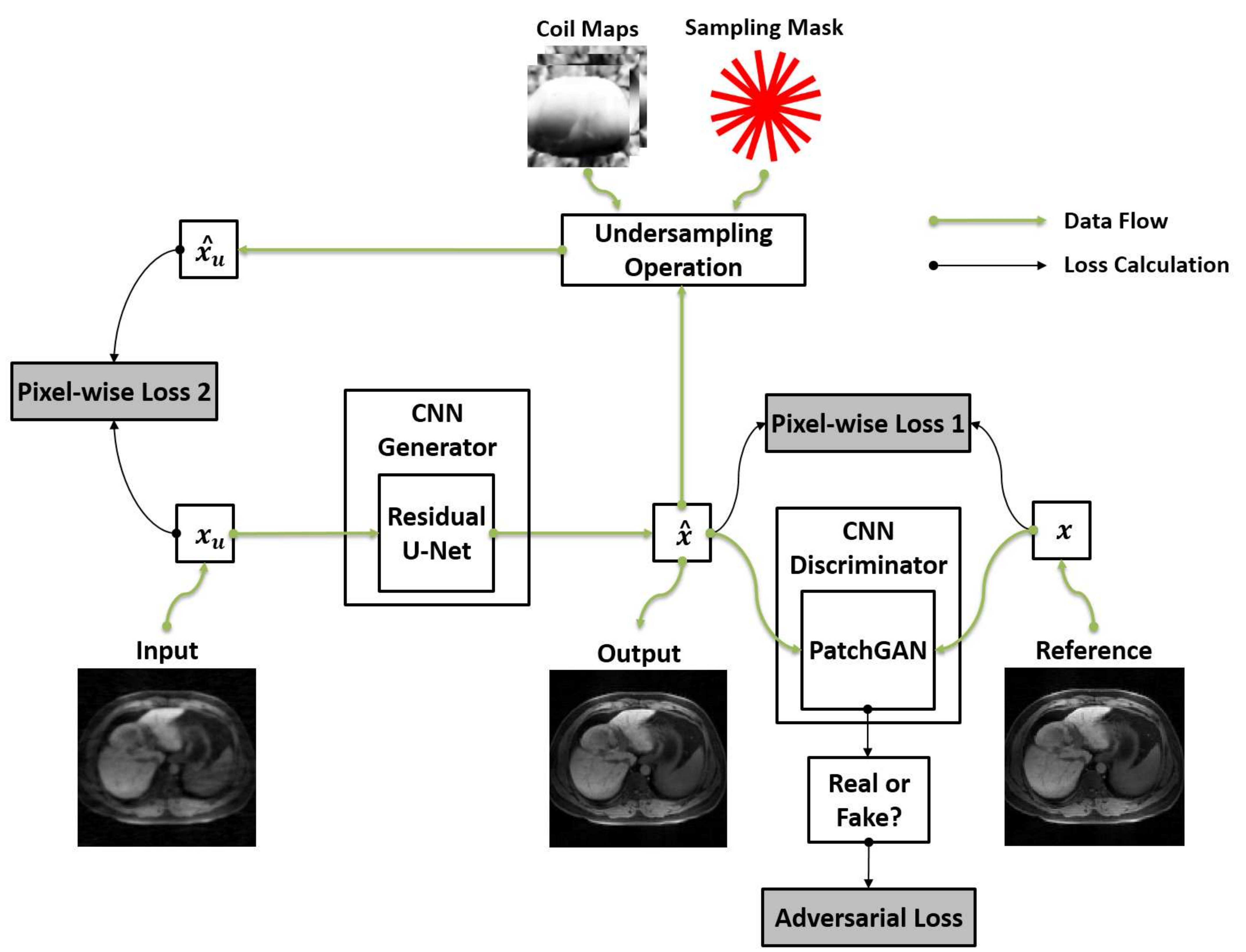}
  \caption{Illustration of the SANTIS reconstruction using a data cycle-consistent adversarial network, which features three loss components. The first loss term (pixel-wise loss 1) removes aliasing artifacts such that the reconstructed image ($\hat x$) looks like their reference ($x$). The second loss term (pixel-wise loss 2) ensures that the reconstructed images from CNN generator (e.g., Residual U-Net) produce undersampled images ($\hat x_u$) matching the acquired \textit{k}-space measurements ($x_u$). The adversarial loss term enforces a high perceptional quality for the reconstructed image to maintain image details and texture using a CNN discriminator (e.g., PatchGAN). The notation in this figure follows the main text description.}
\label{fig1}
\end{figure}

In a practical implementation from recent CNN reconstruction studies (14,29), a cyclic loss can be incorporated to enforce data fidelity in combination with the CNN-regularization as shown in Figure \ref{fig1}. The image reconstruction, expressed in Eq.[6] can then be further reformatted as: 

\begin{equation} \label{eq7}
\hat F = \arg \;\mathop {\min }\limits_F \left( {{\lambda _{loss2}}{\mathbb{E}_{{x_u} \to P({x_u})}}\left[ {{{\left\| {{\Phi _u}F({x_u}) - {x_u}} \right\|}_N}} \right] + {\lambda _{loss1}}{\mathbb{E}_{{x_u} \to P({x_u})}}\left[ {{{\left\| {F({x_u}) - x} \right\|}_N}} \right]} \right)
\end{equation}

where two loss functions were generated including a CNN generator term for CNN-regularization (pixel-wise loss 1: the right term) and a cyclic loss term for data consistency (pixel-wise loss 2: the left term).  ${{\Phi _u} = {C^H}{\Im ^H}U\Im C}$ is the undersampling encoding operator and ${\Phi _u}\hat x = {C^H}{\Im ^H}U\Im C\hat x \to {\hat x_u}$  generates undersampled images using the undersampling mask and coil sensitivities. ${{\mathbb{E}_{{x_u} \to P({x_u})}}\left[  \cdot  \right]}$  is the expectation of a probability function given ${x_u}$ belongs to the data distribution of the undersampled image domain  $P(x_u)$ . $\lambda_{loss1}$ and $\lambda_{loss2}$ are corresponding weighting parameters controlling the balance between the two loss terms. More specifically, the first loss term (pixel-wise loss 1) removes aliasing artifacts such that the reconstructed image looks like their reference, while the second loss term (pixel-wise loss 2) ensures that the reconstructed images from CNN generator produce undersampled images matching the acquired \textit{k}-space measurements. 

\subsection{Adversarial Loss}

Typically, a $l_1$ norm or a $l_2$ norm is often chosen for the pixel-wise loss terms (loss 1 and 2) of Eq.\ref{eq7} in most studies. While either of these two norms can ensure good reconstruction performance in suppressing noise and artifacts, such a reconstruction scheme has been demonstrated to be suboptimal for restoring undersampled images with a ‘natural’ appearance as in corresponding fully sampled images(14,15).  Several studies have shown that using CNN mapping only with pixel-wise losses can result in image blurring, detail loss and degraded perceptional quality (14,15). Therefore, Generative Adversarial Network (GAN) has been recently proposed to overcome the drawbacks of $l_1$ or $l_2$ norm, where an adversarial loss is used to promote image texture preservation in image restoration (30). In the context of image reconstruction as shown in Eq.\ref{eq7}, an adversarial loss can be incorporated into the CNN generator as an additional loss term to promote texture preservation as shown in Figure \ref{fig1}. In other words, in addition to the CNN generator function $F$, another multiple layer CNN discriminator ${D:{f_{dis}}\left( {x|\delta } \right) \to 1}$  conditioned on network parameters $\delta$  is designed to distinguish reconstructed images versus corresponding reference images in the fully sampled image domain $P(x)$ . Mathematically, this discriminator  $D(x)$ outputs a scalar representing the probability that $x$  comes from the fully sampled image domain rather than the CNN generator output $F(x_u)$ . The adversarial loss is formatted as: 

\begin{equation} \label{eq8}
{L_{gan}}(F,D) = {\mathbb{E}_{x \to P(x)}}\left[ {\log D(x)} \right] + {\mathbb{E}_{{x_u} \to P({x_u})}}\left[ {\log (1 - D(F({x_u})))} \right]
\end{equation}

\subsection{Data Cycle-Consistent Adversarial Network}

Following all loss terms mentioned above, a Data Cycle-Consistent Adversarial Network can be formulated for reconstructing undersampled MR images using a multi-loss joint training model as shown in Figure \ref{fig1}. This model consists of a CNN generator that generates a pixel-wise loss term (loss 1) and an adversarial loss term as CNN regularization, and an additional cyclic loss path that creates another pixel-wise loss term (loss 2) for enforcing data fidelity. Therefore, a full objective function can be written as: 

\begin{multline} \label{eq9}
\begin{array}{l}
{L_{full}} = {L_{cyc}} + {L_{gan}} \\
\hspace{2.5em} ={\lambda _{loss2}}{\mathbb{E}_{{x_u} \to P({x_u})}}\left[ {{{\left\| {{\Phi _u}F({x_u}) - {x_u}} \right\|}_N}} \right] + {\lambda _{loss1}}{\mathbb{E}_{{x_u} \to P({x_u})}}\left[ {{{\left\| {F({x_u}) - x} \right\|}_N}} \right]\\
\hspace{3.5em} + {\lambda _{gan}}\left( {{\mathbb{E}_{x \to P(x)}}\left[ {\log D(x)} \right] + {\mathbb{E}_{{x_u} \to P({x_u})}}\left[ {\log (1 - D(F({x_u})))} \right]} \right)
\end{array}
\end{multline}

Here, $\lambda _{gan}$ is a weighting factor to control the contribution of the GAN regularization to the reconstruction. The overall aim of the training is to solve the full objective function in a two-player minimax fashion (30) as: 

\begin{equation} \label{eq10}
\hat F,\hat D = \arg \mathop {\min }\limits_F \mathop {\max }\limits_D {L_{full}}\left( {F,D} \right)
\end{equation}

Namely, $F$ aims to minimize this objective function against the adversary $D$  that seeks to maximize it. In other words, $F$  tries to reconstruct undersampled images into the ones that look similar to fully sampled images, while $D$ aims to distinguish reconstructed images from real fully sampled images. As demonstrated in the studies for natural image-to-image translation (30–32), successful training using such a competing scheme can result in an optimal CNN generator capable of generating artificial images indistinguishable from the original images. Mathematically, this enforces the reconstructed images to reside in the same distribution of the fully sampled images in a high-dimensional manifold (30). It should be noted that optimizing Eq.\ref{eq10} aims to search for optimal network parameters $\hat \theta$  for  ${F:{f_{gen}}\left( {{x_u}|\theta } \right) \to \hat x}$ during the training phase, so that once the training is completed and satisfied, $\hat \theta$ can be fixed for the CNN generator and directly applied to reconstruct a new undersampled image by using the CNN mapping  $F$ to the undersampled image as: 

\begin{equation} \label{eq11}
{x_{cnn}} = {f_{gen}}({x_u}|\hat \theta ),{x_u} \in P({x_u})
\end{equation}

\subsection{Sampling-Augmented Data Cycle-Consistent Adversarial Network}

To strengthen the robustness of the trained network towards sampling pattern discrepancy, the training of Eq.\ref{eq10} can be further augmented in a manner where   was generated to have different artifact patterns at each training iteration. Once the training is successful, the trained network is capable of learning a variety of aliasing artifact patterns and can then be generalized towards removing a broad range of artifacts. This can be achieved by varying the undersampling mask for generating each training data pair throughout the entire training iterations. Eq.\ref{eq9} is then updated as: 

\begin{equation} \label{eq12}
\begin{array}{l}
{L_{full}} = {L_{cyc}} + {L_{gan}} \\
\hspace{2.5em}= {\lambda _{loss1}}{\mathbb{E}_{{x_{u,ij}} \to P({x_u})}}\left[ {{{\left\| {{\Phi _{u,ij}}F({x_{u,ij}}) - {x_{u,ij}}} \right\|}_N}} \right] + {\lambda _{loss2}}{\mathbb{E}_{{x_{u,ij}} \to P({x_u})}}\left[ {{{\left\| {F({x_{u.ij}}) - x} \right\|}_N}} \right]\\
\hspace{3.5em} + {\lambda _{gan}}\left( {{\mathbb{E}_{x \to P(x)}}\left[ {\log D(x)} \right] + {\mathbb{E}_{{x_{u,ij}} \to P({x_u})}}\left[ {\log (1 - D(F({x_{u,ij}})))} \right]} \right)
\end{array}
\end{equation}

where $i$  denotes the  $i^{th}$ image in the training dataset and $j$  denotes the $j^{th}$  training iteration. In other words,  $x_u$ has a different aliasing artifact in each image of the dataset ($x_{u,i}$) and the artifact is also changing for different iterations ($x_{u,ij}$). Accordingly, the undersampling operator ${{\Phi _{u,ij}}}$ is rewritten as: 

\begin{equation} \label{eq13}
\begin{array}{l}
{\Phi _{u.ij}} = {C^H}{\Im ^H}{U_{ij}}\Im C\\
{\Phi _{u.ij}}\hat x = {C^H}{\Im ^H}{U_{ij}}\Im C\hat x \to {{\hat x}_{u.ij}}
\end{array}
\end{equation}

to reflect the variation of undersampling patterns  $U_{ij}$ during training. 

\section{METHODS}
\subsection{In-Vivo Image Datasets}

This study was performed in compliance with the Health Insurance Portability and Accountability Act (HIPPA) regulations and was approved by local Institutional Review Boards (IRB). The knee datasets were acquired from a total of 30 patients with knee pain and discomfort undergoing a clinical knee MR examination, with a waiver of written informed consent. Data were acquired on a 3.0T scanner (Signa Premier, GE Healthcare, USA) equipped with an 18-element knee coil array using a clinical 2D coronal proton density-weighted fast spin-echo (Cor-PD-FSE) sequence. Relevant imaging parameters included: matrix size = 420$\times$448, FOV = 15$\times$15 cm\textsuperscript{2}, TR/TE = 2500/20ms, flip angle = 111 degree, echo train length = 5, slice thickness = 3mm, number of slices = 30-38. 

The liver datasets were acquired in 63 patients who were referred for a clinical liver MR examination. For each patient, data acquisition was added at the end of the scan (approximately 20 minutes after injection of Gd-EOB-DTPA) on a 3.0T scanner (TimTrio, Siemens Healthineers, Germany) equipped with a 12-element body matrix coil, with a written, informed consent obtained before the study. Data were acquired using a prototype fat-saturated stack-of-stars golden-angle radial sequence (33) with the following imaging parameters: matrix size = 256$\times$256, FOV = 33$\times$33 cm\textsuperscript{2}, TR/TE = 3.40/1.68ms, flip angle = 10 degree, slice thickness = 5mm, number of slices = 44, 80\% partial Fourier along the slice dimension. A total of 1000 spokes were acquired in each partition, resulting in a scan time of 178 seconds. 

In addition, a liver Dynamic Contrast-Enhanced (DCE)-MRI data, previously acquired on a healthy volunteer with written, informed consent was also included in this study. This dataset was acquired during contrast injection using the same golden angle radial sequence with a similar protocol as described above, except longer scan time (190 seconds, 1222 spokes for each partition) to ensure capture of contrast dynamics from a pre-contrast phase to a delayed phase.

\subsection{Implementation of Neural Network}

\begin{figure}[h]
  \centering
  \includegraphics[width=0.9\linewidth]{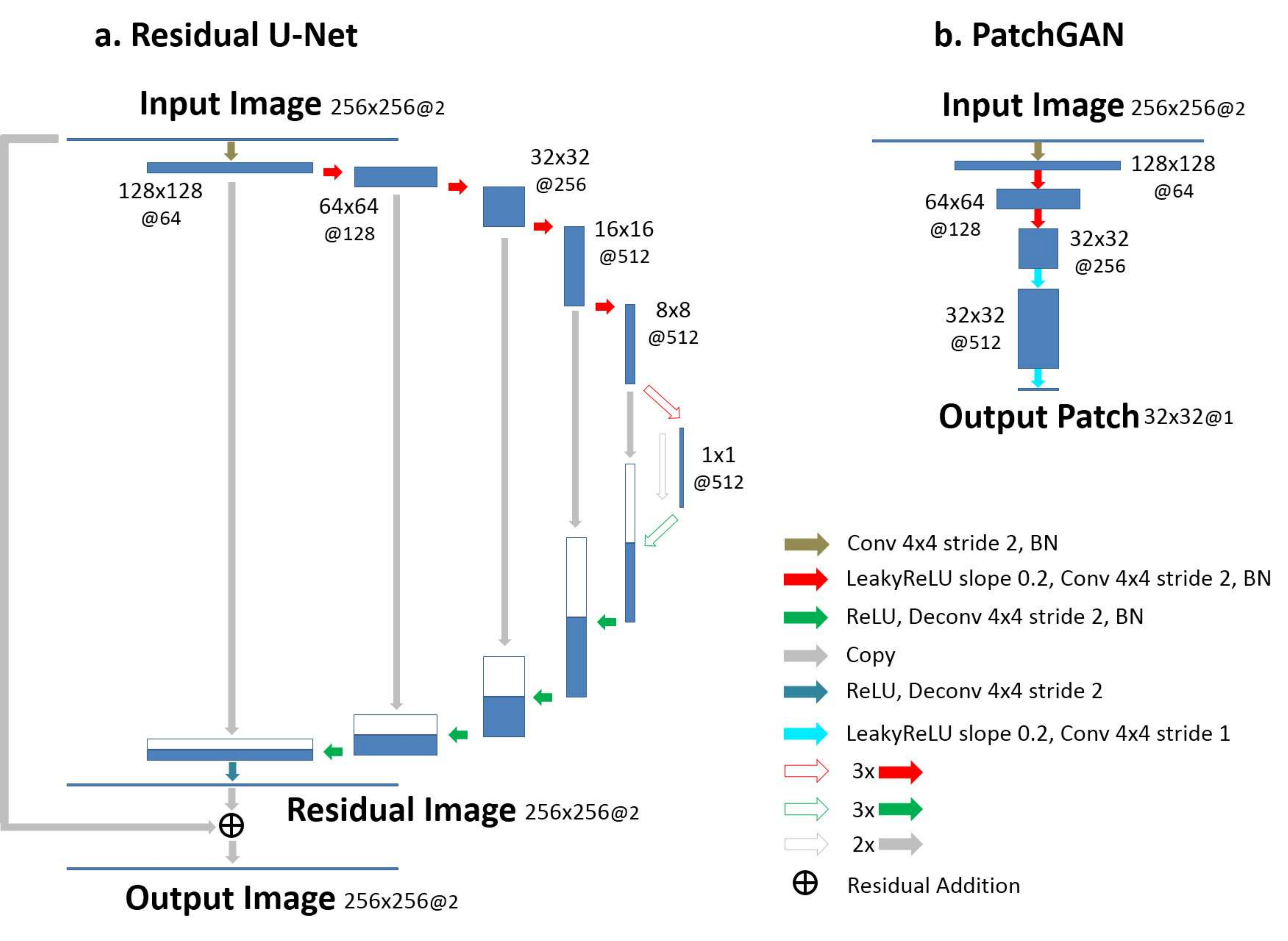}
  \caption{Illustration of the residual U-Net and PatchGAN implemented in SANTIS for end-to-end CNN mapping and adversarial training. The U-Net structure consists of an encoder network and a decoder network with multiple shortcut connection (e.g., concatenation) between them to enhance mapping performance. The abbreviations for the CNN layers include BN for Batch Normalization, ReLU for Rectified Linear Unit activation, Conv for 2D convolution, and Deconv for 2D deconvolution. The parameters for the convolution layers are labeled in the figure as image size @ the number of 2D filters.}
\label{figS1}
\end{figure}

A U-Net architecture (Figure \ref{figS1}) with residual learning was adapted from previous image-to-image translation studies (6,32) (https://github.com/phillipi/pix2pix) for the CNN generator ($F$) to map undersampled image domain into corresponding fully sample image domain. This U-Net structure is composed of an encoder network and a decoder network with multiple shortcut connections between them (34). The encoder is used to achieve efficient data compression while probing robust and spatial invariant image features of input images. A decoder network with a mirrored structure of the encoder is applied for restoring image features and increasing image resolution using the output of the encoder network. Multiple shortcut connections are incorporated to concatenate entirely feature maps from encoder to decoder to enhance mapping performance. Residual learning is implemented to boost learning robustness and accuracy further (35,36). Such a network structure and its variants have shown promising results for image-to-image translation in many recent deep learning studies (6,34–36). For the adversarial configuration, a network architecture developed in PatchGAN (32) was used for the discriminator network ($D$), which aims to differentiate real against artificial images in the adversarial process using image patch-based assessment on input images. An illustration of the applied PatchGAN is also shown in Figure \ref{figS1}. Such a patch-level discriminator architecture has advantages of efficient network training with fewer network parameters and has demonstrated favorable results at differentiating real versus artificial images in many recent GAN studies (6,32,37,38).

\subsection{Network Training}

In the current framework, the complex MR image $x \in {{\mathbb{C}}^M}$ is converted into a new representation $x \in {{\mathbb{R}}^{2M}}$  so that the real and imaginary components are treated as two individual image channels for the network input (14,17). Among all the 30 knee datasets, 22 data were randomly selected and were used for training the deep learning network, and the remaining eight patients were used for evaluation. Similarly, 55 liver data were randomly selected from all the 63 datasets for training, and the remaining eight patients were used for evaluation.

The network weights for the training were initialized using the initialization scheme as described in (39) and were updated using Adam algorithm (40) with a fixed learning rate of 0.0002. The training was performed in a mini-batch manner with 3 image slices in a single mini-batch (6). The $l_1$ norm was applied for the pixel-wise loss terms in Eq.\ref{eq12} to promote image sharpness following recent GAN studies (15). Total iteration steps corresponding to 200 epochs were carried out to ensure convergence of the training process. During each iteration, a two-step training strategy was applied where the CNN generator (F) and adversarial discriminator (D) were updated separately in an alternating manner (6). After qualitatively evaluation with a broad range of combinations, the parameters for the weight factors in the full objective function of Eq.\ref{eq12} were empirically selected as $\lambda_{loss1}$= 10,  $\lambda_{loss2}$= 10 and  $\lambda_{gan}$=0.1 for our study. 

All the proposed algorithms were implemented in Python language (v2.7, Python Software Foundation, Wilmington, DE). The CNNs were designed using the Keras package (41) running Tensorflow computing backend (42) on a 64-bit Ubuntu Linux system. All the training and evaluation were performed on a computer server with an Intel Xeon W3520 quad-core CPU, 32GB RAM, and one Nvidia GeForce GTX 1080Ti graphic card with a total of 3584 CUDA cores and 11GB GDDR5 RAM.

\subsection{Fixed Training vs. Sampling-Augmented Training}

\begin{figure}[h]
  \centering
  \includegraphics[width=0.8\linewidth]{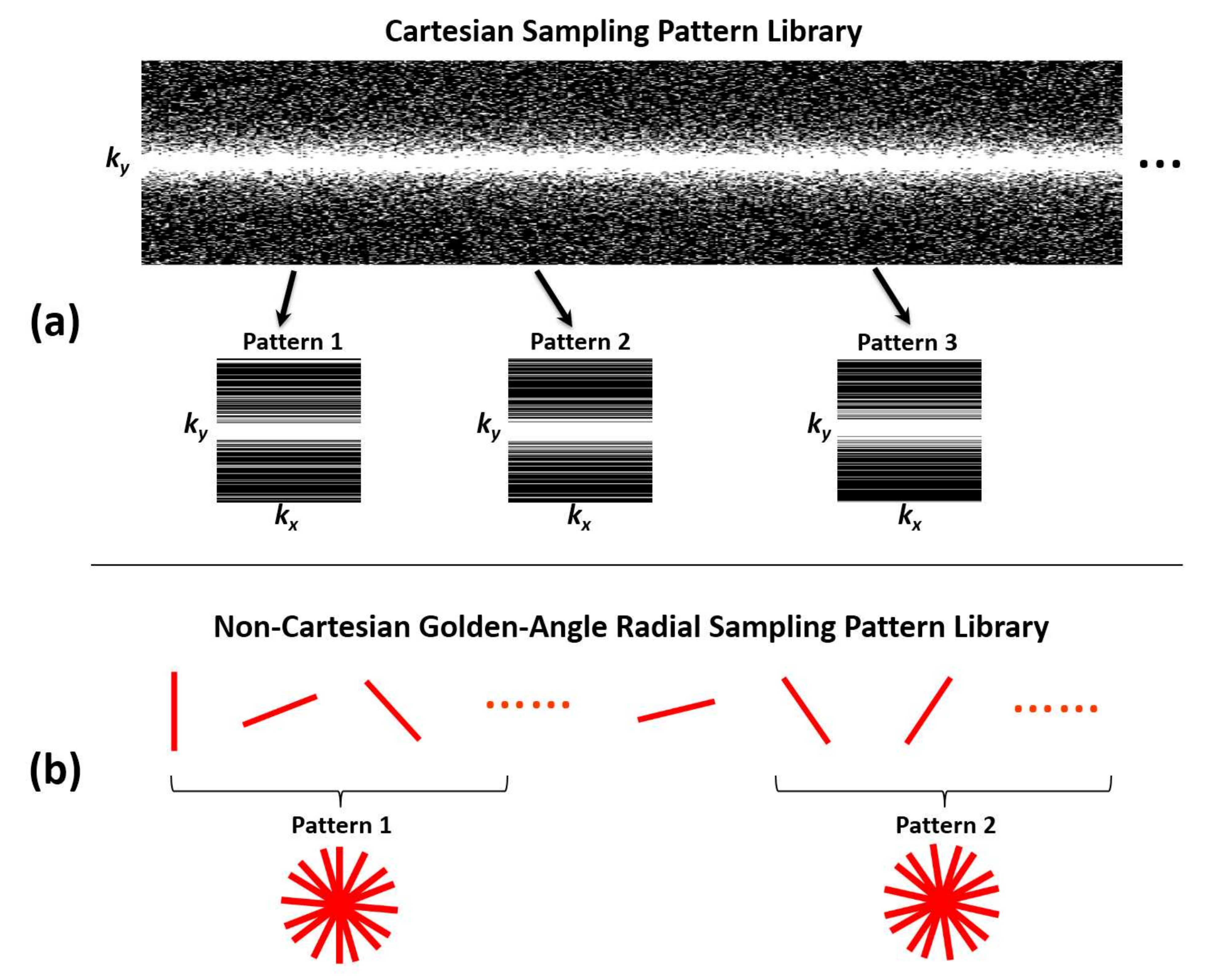}
  \caption{Schematic demonstration of the undersampling patterns used in the study. \textbf{(a)} Examples of the 1D variable-density Cartesian random undersampling patterns used for knee imaging. \textbf{(b)} Example of gold-angle radial undersampling masks used for liver imaging. The undersampling mask was varying for each iteration during the network training to augment the training data for SANTIS framework.}
\label{fig2}
\end{figure}

Two training strategies, including \textbf{1)} the fixed training strategy (referred to as \textbf{CNN-Fix} hereafter) and \textbf{2)} the sampling-augmented training strategy (\textbf{SANTIS}), were investigated in this study for both knee and liver imaging.

A library of 1D variable-density Cartesian random undersampling patterns (25), consisting of 3000 different undersampling masks, were generated for the knee imaging as shown in Figure \ref{fig2}a. For each undersampling mask, the \textit{k}-space center (5\% of the total measurements) was consistently fully sampled. For the training of CNN-Fix, only the first undersampling mask was used throughout the entire training process as outlined in Eq.\ref{eq9}. For the training of SANTIS, in contrast, an undersampling mask was randomly selected from this library for each image in each training iteration, as shown in Eq.\ref{eq12}. This operation ensures that different artifacts can be created for each image at each training iteration so that the network can learn a wide range of undersampling artifact structures during the training. As a pre-processing step, all the knee datasets were compressed into 8 virtual coils to save GPU memory, and corresponding coil sensitivity maps were estimated using the adaptive combination approach (43). To test the acceleration performance, two acceleration rates (R=3 and R=5) were investigated for the knee imaging. 

The training for the liver datasets was performed with the following steps. First, all the datasets were compressed into 4 virtual coils, and a multi-coil 3D liver image (e.g., without coil combination) was reconstructed using a standard gridding algorithm for each dataset. Corresponding coil sensitivity maps were estimated using the adaptive combination approach (43). Second, a library of radial trajectories consisting of 3000 consecutive golden-angle rotations was calculated, from which different undersampled golden-angle patterns can then be generated by combining a subset of consecutive radial trajectories (89 consecutive spokes in this study). Third, using these radial undersampling patterns, corresponding undersampled radial spokes and radial images can then be generated using non-uniform fast Fourier transform (NUFFT) and inverse NUFFT, respective. Fourth, same as that in the knee imaging, the training of CNN-Fix was performed with only the first undersampling pattern (e.g., the 1\textsuperscript{st} to the 89\textsuperscript{th} spokes), while the training of SANTIS was performed by randomly selected a radial undersampling pattern from the trajectory library. Due to the GPU memory limit, the 3D liver datasets were treated as multi-slice 2D images, in a way that both the training and reconstruction were performed slice-by-slice. As a pre-processing step, self-calibrating GRAPPA Operator Gridding (GROG) (44,45) was applied to all the generated reference-undersampled radial data pairs, so that the subsequent training can be performed entirely on a Cartesian grid.

\subsection{Evaluation of Reconstruction Methods}

In our first experiment, the robustness of different training strategies was compared. Specifically, the trained networks from CNN-Fix and SANTIS were applied to reconstruct (the inference step) undersampled images generated with two undersampling masks/patterns, including the one used for CNN-Fix training (denoted as MaskC1 for knee and MaskR1 for liver hereafter) and a newly-generated one not used for training (denoted as MaskC2 for knee and MaskR2 for liver hereafter). It was ensured that both undersampling patterns were not included in the training of SANTIS to fully assess its robustness.

In the second experiment, both CNN-Fix and SANTIS was compared with reconstruction methods previously demonstrated for reconstructing undersampled MR data, including a combination of compressed sensing and parallel imaging, (CS-PI) (46), and a recently proposed Variational Network (VN) (13,28) representing the current state-of-the-art in deep learning-based reconstruction. For the CS-PI, reconstruction was performed slice by lice, and coil sensitivity maps were generated as above. A 6-level Daubechies-4 wavelet transform was selected as the sparsifying transform. The VN approach was implemented using the source code provided by the original developers (https://github.com/VLOGroup/mri-variationalnetwork) with reconstruction parameters that were previously configurated for accelerated knee imaging (13,28). Since the VN approach was only demonstrated for accelerated knee imaging previously, it was only used as a reference for the knee imaging as well in this study. For this comparison, the undersampling patterns MaskC1 and MaskR1 were used for the knee and liver, respectively. 

The reconstruction performance was evaluated using quantitative metrics focusing on different aspects of the reconstruction quality. The normalized Root Mean Squared Error (nRMSE) was used to assess the overall reconstructed image errors. The Structural Similarity Index (SSIM) with a kernel standard deviation of 2.0 was used to estimate the overall image similarity with respect to the reference. The image sharpness was evaluated by calculating the relative reduction of Tenengrad measure (47) between the reconstructed and reference images, where local characteristics of pronounced image edges were explored.

In the third experiment, to test the robustness of SANTIS towards reconstructing images with different contrast, the trained network was also evaluated on the liver DCE-MR dataset with retrospective undersampling. Specifically, the DCE-MR data was reconstructed using RACER-GRASP (48) with a temporal resolution of 15 seconds, resulting in a pre-contrast phase, 3 arterial phases, venous phases at 2 time points and several delayed phases. Undersampled radial images were then generated from these reconstructed contrast phases, using different newly-generated golden-angle radial sampling patterns with 89 consecutive spokes each. 

\section{RESULTS}

\begin{figure}[h]
  \centering
  \includegraphics[width=1\linewidth]{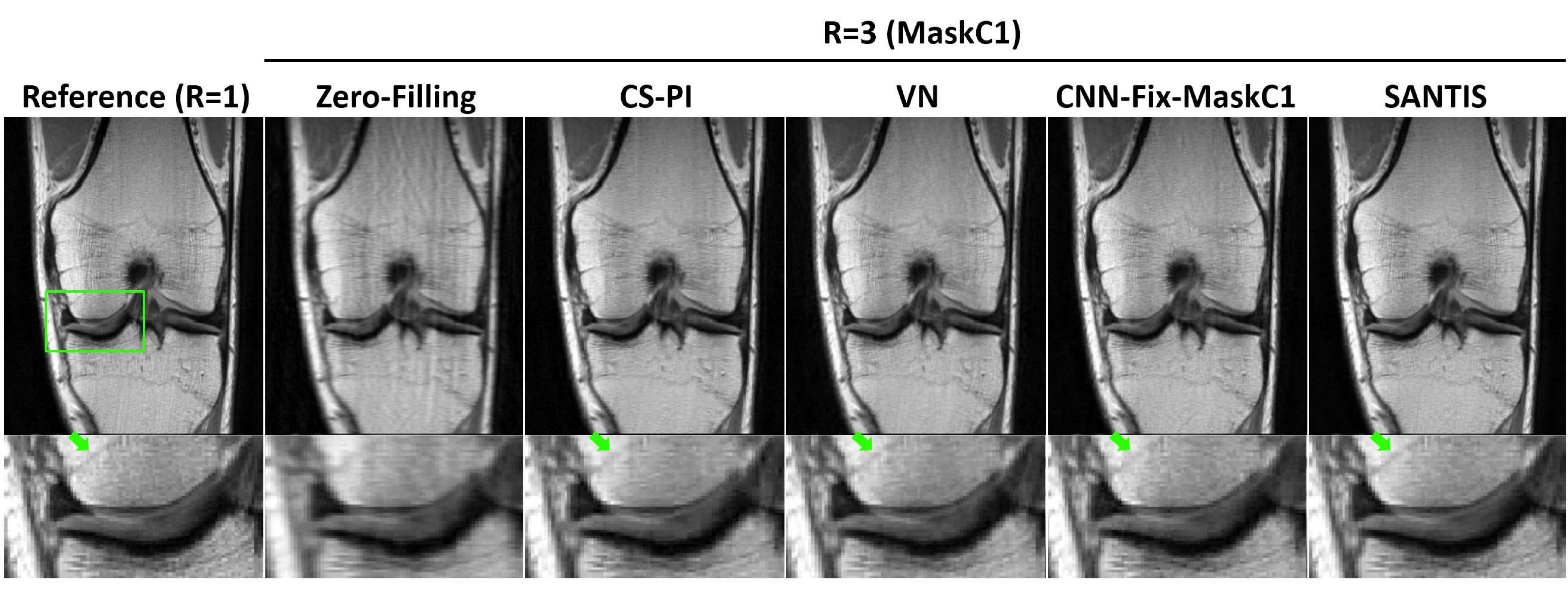}
  \caption{Representative examples of knee images estimated from the different reconstruction methods at R=3, respectively. SANTIS generated a nearly artifact-free image with well-preserved sharpness (green arrows) and texture comparable to the reference.}
\label{fig3}
\end{figure}

Figure \ref{fig3} shows a comparison of different reconstruction methods in one representative knee dataset at R=3 with MaskC1. Compared to the CS-PI, learning-based reconstruction methods (VN, CNN-Fix-MaskC1, and SANTIS) generally provided better restoration from aliasing artifacts in the bone and cartilage, as shown in the zoomed images. SANTIS shows better image restoration than others, resulting in a nearly artifact-free reconstructed image. A sharp texture, indicated by the green arrows, was better preserved in SANTIS compared to other methods and is closely similar to the fully sampled reference. 

\begin{figure}[h]
  \centering
  \includegraphics[width=1\linewidth]{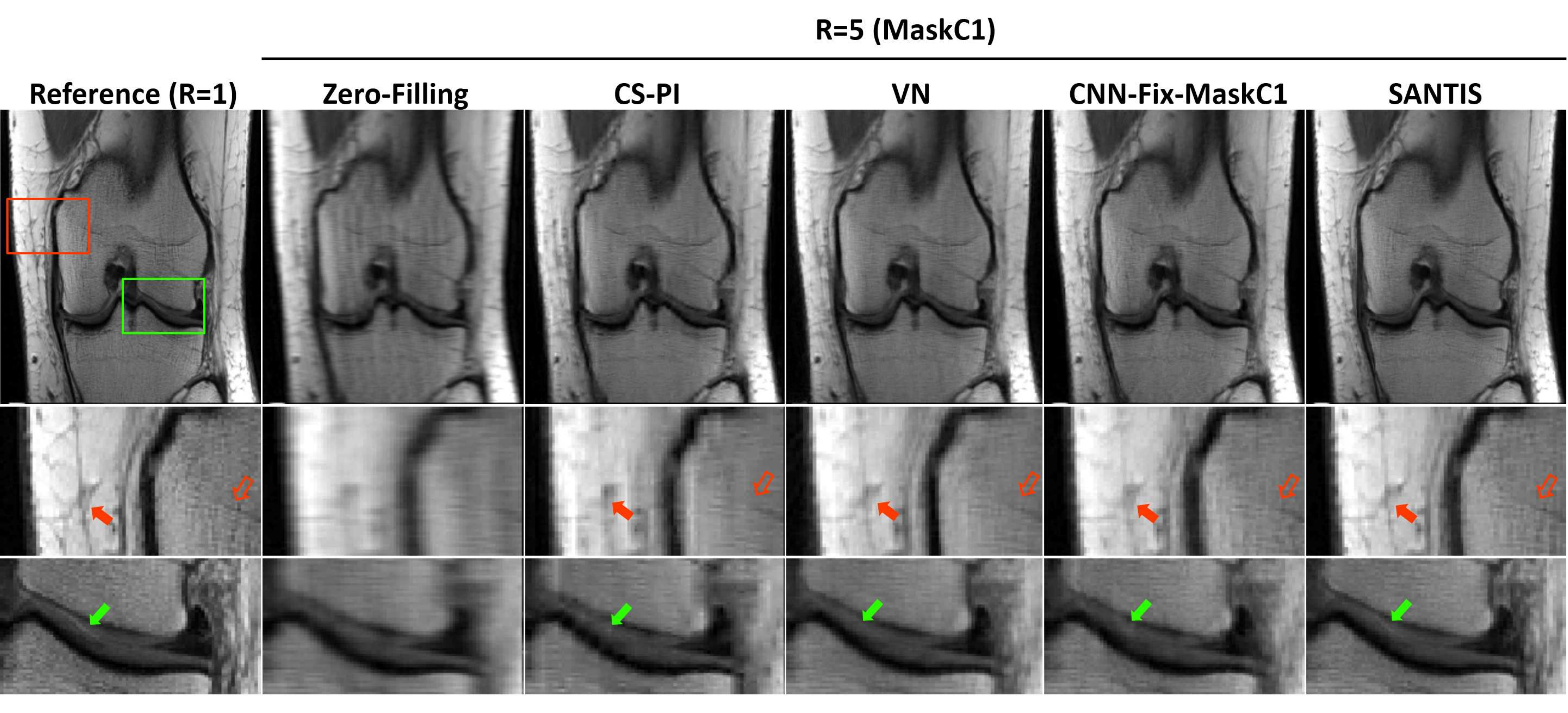}
  \caption{Representative examples of knee images estimated from the different reconstruction methods at R=5, respectively. Compared to other methods, SANTIS generated a better representation for the layer structure of the femoral and tibial cartilage (green arrows) with well-preserved sharpness and image texture (red arrows) comparable to the reference.}
\label{fig4}
\end{figure}

The same comparison at R=5 is shown in Figure \ref{fig4} for another dataset. Similar improvement of image quality can also be observed in SANTIS, with better representation for the layer structure of the femoral and tibial cartilage (green arrows). At this acceleration, CS-PI failed to reconstruct undersampled image faithfully, and both VN and CNN-Fix suffered from notable residual blurring and loss of image details, as shown by the red arrows in the zoomed bone and subcutaneous fat regions.

\begin{figure}[h]
  \centering
  \includegraphics[width=0.8\linewidth]{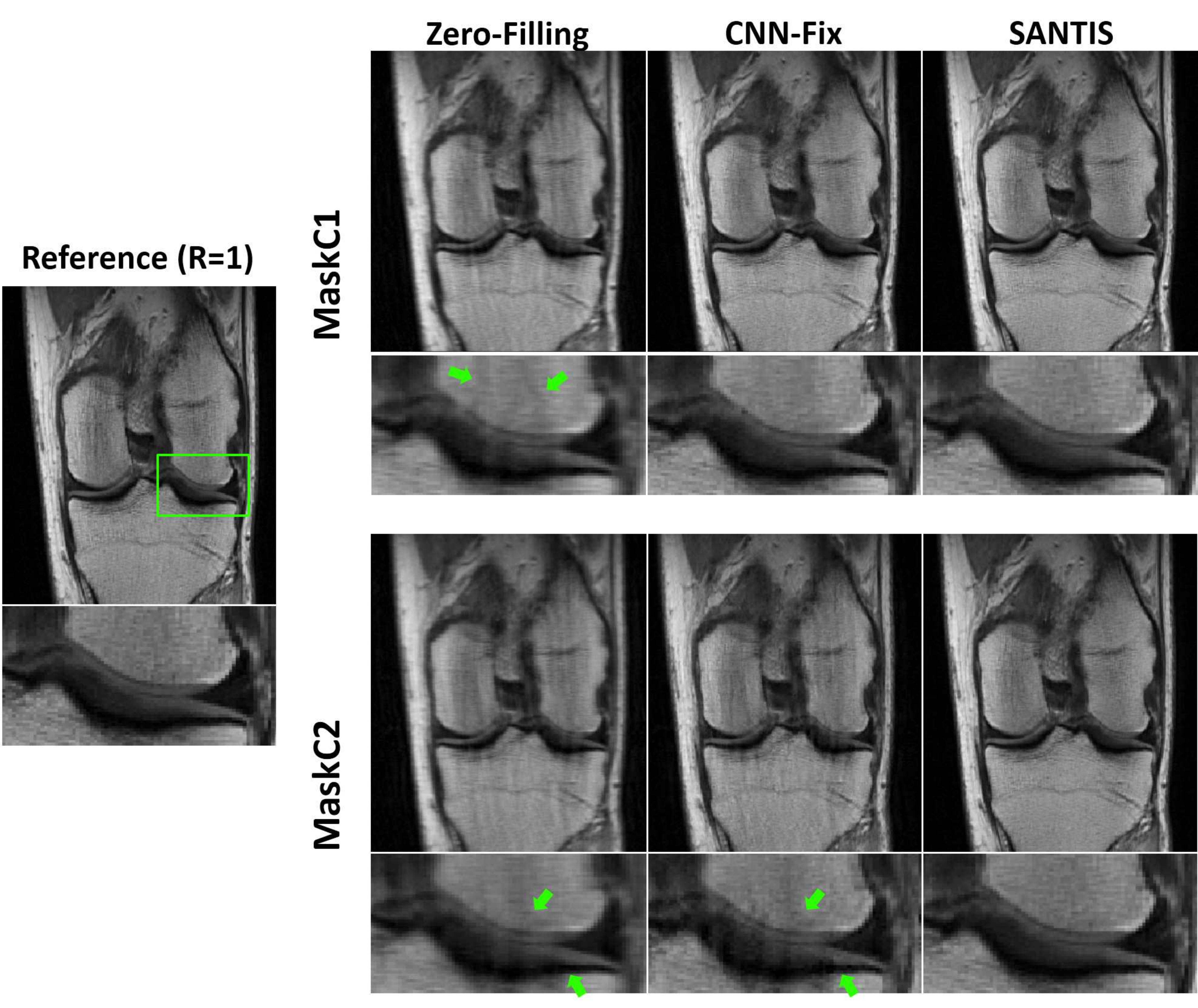}
  \caption{Evaluation of reconstruction robustness at different undersampling masks for MaskC1 and MaskC2. The fixed training CNN-Fix failed to reconstruct MaskC2 with noticeable residual artifacts indicated by the green arrows. SANTIS provided a consistent high-quality reconstruction for both masks which is better than CNN-Fix.}
\label{fig5}
\end{figure}

Figure \ref{fig5} shows knee images comparing CNN-Fix with SANTIS using both MaskC1 (top row) and MaskC2 (bottom row) at R=3. As described in the Methods section, only the MaskC1 was used for the training of CNN-Fix, and both MaskC1 and MaskC2 were not used for the SANTIS training. It is expected that MaskC1 and MaskC2 created different aliasing artifacts, as highlighted by the green arrows in the zero-filling images. While CNN-Fix-MaskC1 shows decent image quality, CNN-Fix-MaskC2 failed to faithfully reconstruct the undersampled image generated with MaskC2, with noticeable residual aliasing artifacts in the zoomed image indicated by the green arrows. In contrast, SANTIS demonstrates robust reconstruction for both MaskC1 and MaskC2, with consistently better visual image quality than CNN-Fix. 

\begin{figure}[h]
  \centering
  \includegraphics[width=1\linewidth]{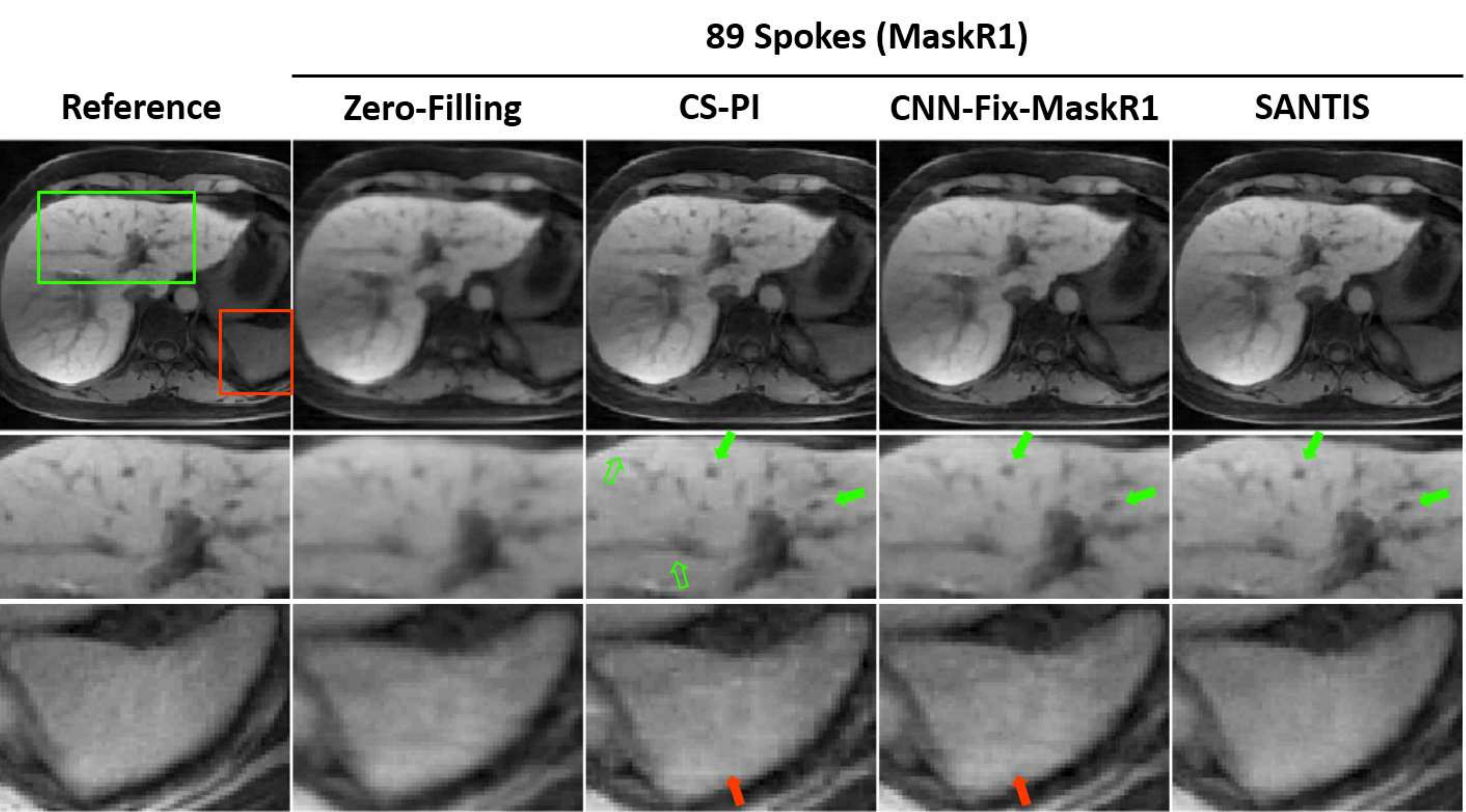}
  \caption{A representative comparison of different reconstruction methods in radial liver imaging with MaskR1. SANTIS achieved the best visual image quality with successful removal of streaking artifacts and with favorable preservation of tissue sharpness and texture. CS-PI shows some pseudo residual structures (hollow green arrows). Meanwhile, CNN-Fix-MaskR1 suffers from noticeable residual blurring as indicated by the green arrows. For both CS-PI and CNN-Fix-MaskR1, residual streaking artifacts can be noted, as highlighted by the red arrows in the zoomed spleen region.}
\label{fig6}
\end{figure}

A representative comparison of different reconstruction methods in radial liver imaging with MaskR1 is shown in Figure \ref{fig6}. Similar to that in the knee imaging, SANTIS achieved the best visual image quality with successful removal of streaking artifacts and with favorable preservation of tissue sharpness and texture. CS-PI shows some pseudo residual structures (hollow green arrows), which were potentially caused by the use of wavelet constraint. Meanwhile, CNN-Fix-MaskR1 suffers from noticeable residual blurring as indicated by the green arrows. For both CS-PI and CNN-Fix-MaskR1, residual streaking artifacts can be noted, as highlighted by the red arrows in the zoomed spleen region. 

\begin{figure}[h]
  \centering
  \includegraphics[width=0.8\linewidth]{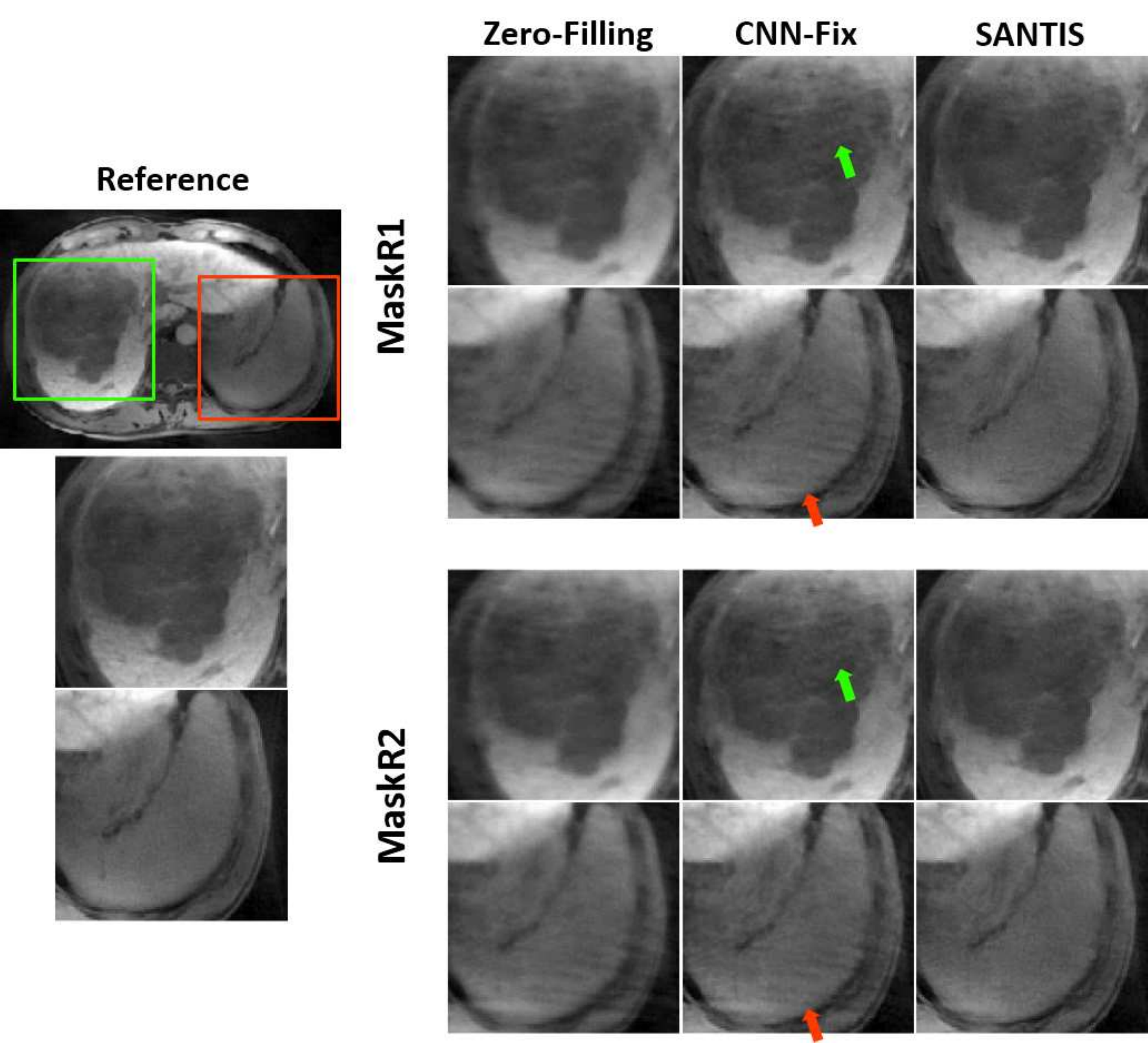}
  \caption{Example of zoomed liver images with a tumor comparing CNN-Fix and SANTIS using both MaskR1 and MaskR2 for inference. The results of both CNN-Fix-MaskR1 and CNN-Fix-MaskR2 suffered from residual artifacts (red arrows), while SANTIS consistently shows improved image quality for both MaskR1 and MaskR2, with a reduced level of residual streaking artifacts and a delineation of the tumor that is closer to the reference.}
\label{fig7}
\end{figure}

\begin{figure}[h]
  \centering
  \includegraphics[width=1\linewidth]{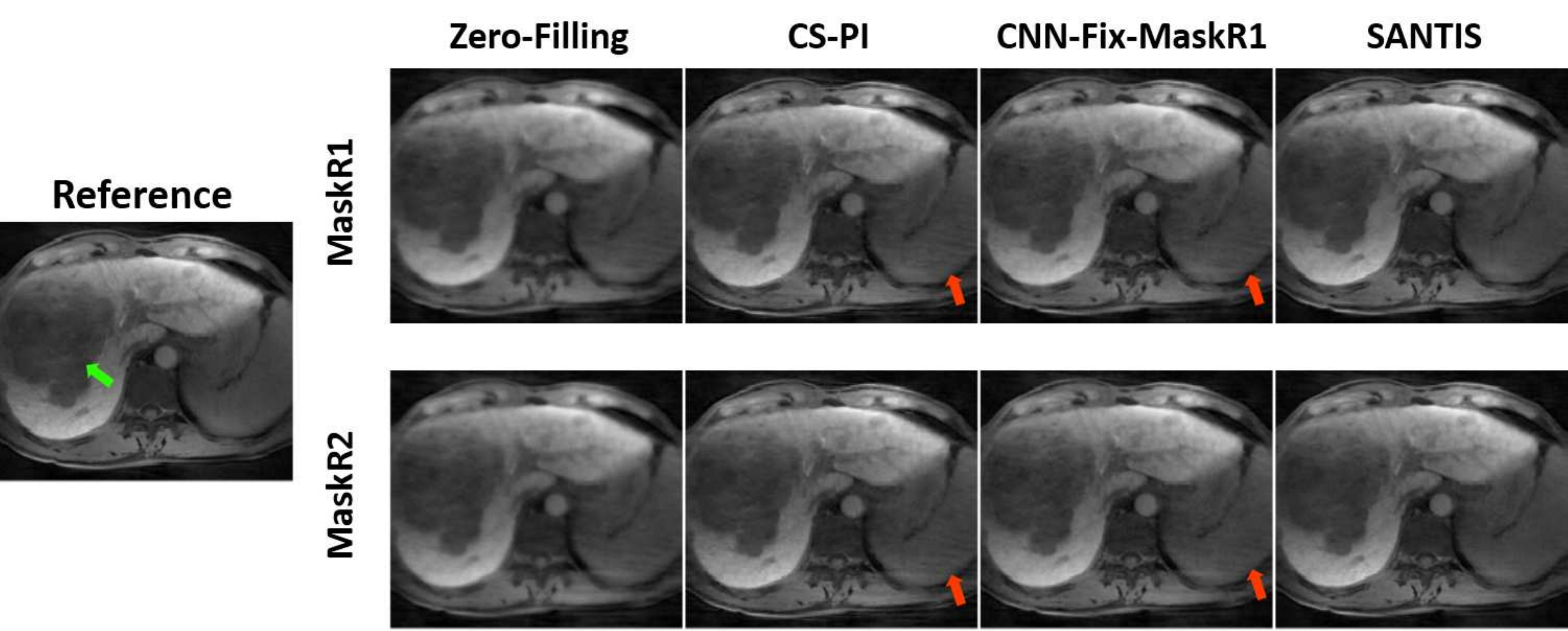}
  \caption{The corresponding full FOV images of Figure \ref{fig7} comparing CNN-Fix and SANTIS using both MaskR1 and MaskR2 for inference.}
\label{figS2}
\end{figure}

Figure \ref{fig7} shows zoomed liver images with a tumor (green box in the reference image) comparing CNN-Fix and SANTIS using both MaskR1 and MaskR2 for inference. The results of both CNN-Fix-MaskR1 and CNN-Fix-MaskR2 suffered from residual artifacts, while SANTIS consistently shows improved image quality for both MaskR1 and MaskR2, with a reduced level of residual streaking artifacts and a delineation of the tumor that is closer to the reference. The corresponding full FOV images of this comparison is shown in Figure \ref{figS2}. 

\begin{table}[t]
  \caption{Quantitative metrics between the reference fully sampled images and the reconstructed images from different reconstruction methods for Cartesian knee imaging.  Results were averaged over the 8 test datasets and represent mean value$\pm$standard deviation. SANTIS achieved the highest and most robust reconstruction performance at both R=3 and 5.}
   \label{tb1}
  \centering
  \begin{tabular}{lllllllll}
    \toprule
    && \multicolumn{4}{c}{R=3} \\
    \cmidrule{3-6}
    Methods  & &  nRMSE(\%)      &  SSIM(\%)  &  Tenengrad(\%)  &  Average Recon Time(s) \\
    \midrule
    ZF 	&  & 9.87$\pm$1.16  & 82.25$\pm$3.77 & 43.75$\pm$2.69  & 0.001  \\
    CS-PI   &    & 6.74$\pm$0.63  & 87.64$\pm$2.22 & 15.66$\pm$1.99  & 2.12 \\
    VN  & & 6.01$\pm$1.12  & 90.21$\pm$2.54  & 10.11$\pm$2.06  & 0.11   \\
    \multirow{2}{*}{CNN-Fix}
    &MaskC1     & 5.67$\pm$0.94  & 90.78$\pm$2.89  & 10.30$\pm$2.34  & 0.06   \\
    &MaskC2    & 6.92$\pm$1.09  & 86.33$\pm$2.91  & 10.67$\pm$2.86  & 0.06   \\
    \multirow{2}{*}{SANTIS}
    &MaskC1     & 5.15$\pm$1.07  & 91.41$\pm$2.82  & 8.67$\pm$2.45  & 0.06  \\
    &MaskC2     & 5.10$\pm$1.02  & 91.96$\pm$2.90  & 8.51$\pm$2.38  & 0.06  \\
    \\
    
    && \multicolumn{4}{c}{R=5} \\
    \cmidrule{3-6}
    ZF 	&  & 13.58$\pm$1.75  & 75.38$\pm$4.92 & 55.19$\pm$3.52  & 0.001  \\
    CS-PI   &    & 10.91$\pm$1.13  & 83.78$\pm$3.80 & 33.01$\pm$2.99  & 2.21  \\
    VN  & & 10.81$\pm$1.67  & 85.11$\pm$3.96  & 20.13$\pm$3.10  & 0.12  \\
    \multirow{2}{*}{CNN-Fix}
    &MaskC1     & 10.48$\pm$1.41  & 85.06$\pm$4.79 & 19.63$\pm$4.10  & 0.06   \\
    &MaskC2    & 11.56$\pm$1.94  & 82.98$\pm$4.82  & 20.59$\pm$4.95  & 0.06    \\
    \multirow{2}{*}{SANTIS}
    &MaskC1     & 9.60$\pm$1.81  & 89.06$\pm$4.97  & 17.29$\pm$3.79  & 0.06  \\
    &MaskC2     & 9.66$\pm$1.40  & 88.66$\pm$4.62  & 17.11$\pm$3.84  & 0.06    \\    
    
    \bottomrule
  \end{tabular}
\end{table}

\begin{table}[t]
  \caption{Quantitative metrics between the reference fully sampled images and the reconstructed images from different reconstruction methods for golden-angle liver imaging.  Results were averaged over the 8 test datasets and represent mean value$\pm$standard deviation. SANTIS achieved the highest and most robust reconstruction performance.}
   \label{tb2}
  \centering
  \begin{tabular}{lllllllll}
    \toprule
    && \multicolumn{4}{c}{89 spokes}\\
    \cmidrule{3-6}
    Methods   & & nRMSE(\%)      &  SSIM(\%)  &  Tenengrad(\%)  &  Average Recon Time(s) \\
    \midrule
    ZF 	&  & 11.69$\pm$1.39  & 85.80$\pm$2.38 & 37.98$\pm$2.92  & 0.001   \\
    CS-PI   &    & 9.50$\pm$1.56  & 87.42$\pm$3.14 & 3.55$\pm$2.75  & 2.32   \\

    \multirow{2}{*}{CNN-Fix}
    &MaskR1     & 9.00$\pm$1.32  & 91.07$\pm$2.27  & 6.16$\pm$2.33  & 0.06   \\
    &MaskR2    & 9.29$\pm$1.32  & 89.80$\pm$2.24  & 6.76$\pm$2.43  & 0.06 \\
    \multirow{2}{*}{SANTIS}
    &MaskR1     & 8.82$\pm$1.31  & 92.66$\pm$2.00  & 2.00$\pm$2.72  & 0.06   \\
    &MaskR2     & 8.78$\pm$1.29  & 92.17$\pm$2.00  & 1.77$\pm$3.19  & 0.06 \\
    \bottomrule
  \end{tabular}
\end{table}

The group-wise quantitative analyses further confirm the qualitative observation in exemplary figures. The comparisons of reconstruction methods using quantitative metrics, including the nRMSE, SSIM and Tenengrad, are summarized in Table \ref{tb1} and Table \ref{tb2} for all the test knee and liver images, respectively. For knee imaging at both R=3 and R=5 (Table \ref{tb1}), learning-based reconstruction methods (VN, CNN-Fix-MaskC1, and SANTIS) all achieved better image quality than CS-PI, as reflected by the increased SSIM and reduced nRMSE and Tenengrad. Among the learning-based reconstructions, SANTIS produced the best image quality quantitatively. It should be noted that the image quality is degraded from CNN-Fix-MaskC1 to CNN-Fix-MaskC2 when an undersampling pattern (MaskC2) that was not included for the training was used for inference, while SANTIS shows more consistent performance for both MaskC1 and MaskC2. For liver imaging (Table \ref{tb2}), CNN-Fix-MaskR1 and SANTIS also achieved better image quality than CS-PI. However, different from Cartesian knee imaging, the results of CNN-Fix-MaskR1 in radial liver imaging is only marginally degraded compared to CNN-Fix-MaskR2, which is likely due to the more incoherent undersampling behavior in the radial sampling. Nevertheless, the performance of SANTIS is still better than CNN-Fix quantitatively for both MaskR1 and MaskR2. Regarding reconstruction time, the learning-based methods enabled increased reconstruction speed compared to CS-PI, and the SANTIS and CNN-Fix provided the fastest image reconstruction at an order of 60ms per image slice.

\begin{figure}[h]
  \centering
  \includegraphics[width=0.7\linewidth]{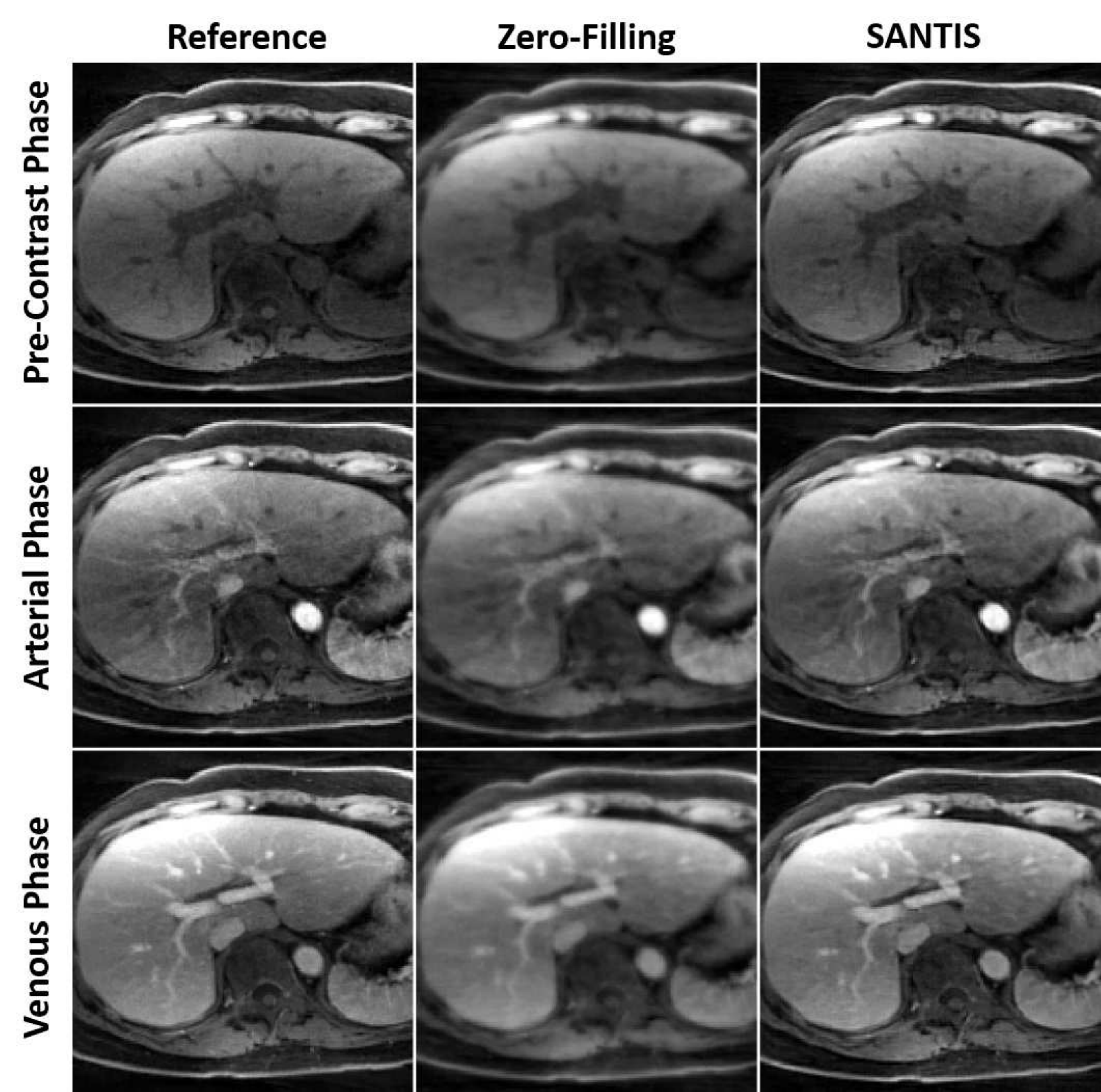}
  \caption{Representative examples demonstrating the performance of SANTIS in reconstructing accelerated liver images at different contrast phases. SANTIS can reconstruct images at a different contrast phase (e.g., pre-contrast phase, arterial phase, or venous phase) without affecting image contrast, and the recovery of image quality from the zero-filling results can be appreciated.}
\label{fig8}
\end{figure}

Finally, an example demonstrating the performance of SANTIS in reconstructing accelerated liver images at different contrast phases is shown in Figure \ref{fig8}. Interestingly, although the SANTIS training was performed using images acquired at a delayed contrast phase, its performance of reconstructing images at a different contrast phase (e.g., pre-contrast phase, arterial phase, or venous phase) is visually plausible without affecting image contrast, and the recovery of image quality from the zero-filling results can be clearly appreciated. 

\section{DISCUSSION}
The main contribution of this work was to propose a novel deep learning-based reconstruction framework called SANTIS for reconstructing accelerated MR images more faithfully with improved robustness against undersampling pattern discrepancy. In addition to a combination of efficient end-to-end CNN mapping, data fidelity enforcement and adversarial training using GAN, SANTIS further incorporates sampling or \textit{k}-space trajectory augmentation with extensive variation of undersampling masks to promote training robustness. The trained network is capable of exploring a complete set of aliasing artifact structures in the training process and thus can be applied to remove artifacts and noises more faithfully. The improved performance of SANTIS was demonstrated for accelerated knee images using a Cartesian trajectory and for accelerated liver images using a golden-angle radial trajectory. Different quantitative metrics were used to assess its performance against fully sampled artifact-free images and different reference reconstruction approaches. 

Despite rapid growth of interest in learning-based reconstruction, the first efforts were focused on the improvement of reconstruction accuracy, applicability, and efficiency (13–21). The robustness of the method, in spite of its importance, has been less studied and remains an essential component in the emerging era of deep learning reconstruction. Here, a general open question is whether a trained network performing well on one set of image dataset can be well-generalized towards reconstructing a different set of datasets with varying image contrast, noise level, imaging hardware or acquisition protocols. A pilot study by Knoll et al. has investigated this question using a variational network (28). Although their findings were insightful, a practical solution was not provided to address this issue. The SANTIS framework in this study represents a first attempt to resolve such kind of challenge in deep learning-based reconstruction. Although we are only aiming at the robustness of acquisition trajectories, the framework might be potentially extended to handle different types of image contrast and even noise levels. 
The performance of SANTIS was investigated using both Cartesian and radial trajectories. It is expected that Cartesian sampling is more sensitive to the discrepancy of undersampling pattern because undersampling is typically performed along the phase encoding dimension only, resulting in less incoherent aliasing structures (Figure \ref{fig5}). Our results have suggested that the improvement of reconstruction performance with SANTIS is evident in Cartesian sampling. Different from Cartesian sampling, radial sampling usually resulted in more benign/incoherent artifacts with a non-specific artifact pattern that can be identified with fewer efforts in the learning-based reconstruction (Figure \ref{fig7}). Nevertheless, SANTIS outperformed CNN-Fix both for Cartesian and for radial sampling patterns regardless of the undersampling masks used during inference. This improvement can be justified in two-fold. On the one hand, the variation of undersampling patterns in SANTIS creates abundant image features which maximize the artifact appeared in the training process and thus helps to better characterize undersampling artifacts for improved image restoration. On the other hand, the variation of the undersampling pattern may introduce additional incoherence, facilitating better separation of undersampled artifacts from real image structure. This resembles the effect of incoherent sampling in compressed sensing (25) and the randomized acquisition in MR fingerprinting (49). The idea of the sampling-augmented training in SANTIS can be treated as a general framework and is not limited to our implemented network, and it is expected that other existing deep learning-based reconstruction techniques may also benefit from such a training strategy for improved reconstruction robustness.

In addition to evaluating accelerated images at the same image contrast, we have also extended the SANTIS framework for reconstructing liver images at different contrast phases (Figure \ref{fig8}). Our first attempt, despite a proof-of-concept trial, has demonstrated encouraging results to reconstruct a rapid contrast wash-in phases (e.g., arterial phase) using image features learned from a delayed phase, suggesting a potential solution of applying deep learning to reconstruct accelerated DCE-MRI data using transfer learning. It was observed in this example that SANTIS is capable of learning essential image features that are uniquely dependent on the undersampling pattern and this is less affected by the change of image contrast caused by contrast injection. In addition to the sampling-augmented training, such performance is also partly due to the use of a combination of residual learning and data fidelity loss. The data fidelity ensures that the reconstructed image matches acquired undersampled measurement, while the residual learning strategy aims to directly subtract image artifacts from undersampled input images, which is advantageous in identifying image artifacts without affecting image contrast and anatomy (35,36). However, further study on a large of datasets is warranted to thoroughly investigate the applicability and utility of this intriguing feature in the SANTIS framework.

The application of adversarial training has been proven to be quite successful and helpful in perseveration of perceptional image structures and texture in reconstructed images, as shown in (14,15). In contrast to the reconstruction methods using solely pixel-wise losses (e.g., VN) that might alter underline image texture and noise spectrum, the reconstructed images using adversarial loss is more advantageous for maintaining a visually plausible image appearance to human readers. The adversarial learning ensures that the restored images fall into the same data distribution of the reference fully sampled images in a transformed high dimensional image manifold (30,50), while the data cycle consistency enforcing data fidelity further prevents the degeneration of the adversarial process from generating hallucinated image features (31). The incorporation of these two components in SANTIS imposes both high-quality image feature learning and information consistency to encourage efficient and accurate CNN end-to-end mapping. However, the training of the adversarial loss is known as a challenge due to the competing mechanism of CNN generator and discriminator. Many newly-developed CNN structures and advanced GAN training strategies, such as LS-GAN (51), Wasserstein GAN (52), conditional GAN (53) and progressive GAN (54) etc, might be further explored in a future study to improve the performance of SANTIS.

There are several limitations in this work that warranty discussion. First, although the SANTIS training was performed using a library of different undersampling patterns, they were generated to have the same 1D variable-density distribution in our Cartesian knee imaging. The robustness of the trained network for reconstructing a completely different sampling scheme (e.g., regular undersampling) remains to be explored. However, it is straightforward to extend the SANTIS training to incorporate different undersampling schemes to include different sampling distribution and different acceleration rates, so that the generalization of the network can be further improved. Second, this study did not analyze the speed of training convergence between CNN-Fix and SANTIS. For both networks, the training was performed with a sufficient number of epochs to ensure training convergence. However, further studies to evaluate the influence of sampling-augmented training on convergence speed would be necessary. Third, the reconstruction parameters, as shown in Eq.\ref{eq12}, were empirically determined in this work. The weight of the GAN regularization, in particular, is curial to ensure good reconstruction performance (14,15). However, selection of reconstruction parameters is not only a challenge for our work, but remains a general open question in constrained image reconstruction. Finally, due to a requirement for a large number of training images, we only evaluated our technique on a few datasets and one DCE dataset. Additional evaluation of the SANTIS framework on a large number of clinical datasets would still be necessary to further assess the value and performance of our technique towards routine clinical use.

\section{CONCLUSIONS}
The framework of SANTIS represents a novel concept for deep learning-based image reconstruction. With sampling or \textit{k}-space trajectory-augmented training by extensively varying undersampling pattern, the framework can recognize different aliasing artifact structures and can be used to remove image artifacts robustly. The concept of SANTIS can be particularly useful towards improving the robustness of deep learning-based image reconstruction against discrepancy between the training and evaluation phases. 

\section{ACKNOWLEDGMENT}
The authors thank the technologists at the UW Hospital in Madison, USA for acquiring the knee datasets with institutional IRB approval. The authors also thank the technologists at the Southwest Hospital in Chongqing, China for acquiring the golden-angle radial liver data with institutional IRB approval.

\section*{REFERENCES}
\medskip
\small

1. Lakhani P, Sundaram B. Deep Learning at Chest Radiography: Automated Classification of Pulmonary Tuberculosis by Using Convolutional Neural Networks. Radiology [Internet] 2017;284:574–582. doi: 10.1148/radiol.2017162326.

2. De Fauw J, Ledsam JR, Romera-Paredes B, et al. Clinically applicable deep learning for diagnosis and referral in retinal disease. Nat. Med. [Internet] 2018:1. doi: 10.1038/s41591-018-0107-6.

3. Esteva A, Kuprel B, Novoa RA, Ko J, Swetter SM, Blau HM, Thrun S. Dermatologist-level classification of skin cancer with deep neural networks. Nature [Internet] 2017;542:115–118. doi: 10.1038/nature21056.

4. Liu F, Zhou Z, Jang H, Samsonov A, Zhao G, Kijowski R. Deep Convolutional Neural Network and 3D Deformable Approach for Tissue Segmentation in Musculoskeletal Magnetic Resonance Imaging. Magn. Reson. Med. [Internet] 2017:DOI: 10.1002/mrm.26841. doi: 10.1002/mrm.26841.

5. Zhou Z, Zhao G, Kijowski R, Liu F. Deep Convolutional Neural Network for Segmentation of Knee Joint Anatomy. Magn. Reson. Med. 2018:doi:10.1002/mrm.27229.

6. Liu F. SUSAN: Segment Unannotated image Structure using Adversarial Network. Magn Reson Med 2018:DOI: 10.1002/mrm.27627.

7. Norman B, Pedoia V, Majumdar S. Use of 2D U-Net Convolutional Neural Networks for Automated Cartilage and Meniscus Segmentation of Knee MR Imaging Data to Determine Relaxometry and Morphometry. Radiology [Internet] 2018:172322. doi: 10.1148/radiol.2018172322.

8. Zhao G, Liu F, Oler JA, Meyerand ME, Kalin NH, Birn RM. Bayesian convolutional neural network based MRI brain extraction on nonhuman primates. Neuroimage [Internet] 2018;175:32–44. doi: 10.1016/j.neuroimage.2018.03.065.

9. Liu F, Zhou Z, Samsonov A, Blankenbaker D, Larison W, Kanarek A, Lian K, Kambhampati S, Kijowski R. Deep Learning Approach for Evaluating Knee MR Images: Achieving High Diagnostic Performance for Cartilage Lesion Detection. Radiology [Internet] 2018:172986. doi: 10.1148/radiol.2018172986.

10. Cicero M, Bilbily A, Colak E, Dowdell T, Gray B, Perampaladas K, Barfett J. Training and Validating a Deep Convolutional Neural Network for Computer-Aided Detection and Classification of Abnormalities on Frontal Chest Radiographs. Invest. Radiol. [Internet] 2017;52:281–287. doi: 10.1097/RLI.0000000000000341.

11. Anthimopoulos M, Christodoulidis S, Ebner L, Christe A, Mougiakakou S. Lung Pattern Classification for Interstitial Lung Diseases Using a Deep Convolutional Neural Network. IEEE Trans. Med. Imaging [Internet] 2016;35:1207–1216. doi: 10.1109/TMI.2016.2535865.

12. Kooi T, Litjens G, van Ginneken B, Gubern-Mérida A, Sánchez CI, Mann R, den Heeten A, Karssemeijer N. Large scale deep learning for computer aided detection of mammographic lesions. Med. Image Anal. [Internet] 2017;35:303–312. doi: 10.1016/j.media.2016.07.007.

13. Hammernik K, Klatzer T, Kobler E, Recht MP, Sodickson DK, Pock T, Knoll F. Learning a Variational Network for Reconstruction of Accelerated MRI Data. Magn. Reson. Med. [Internet] 2017;79:3055–3071. doi: 10.1002/mrm.26977.

14. Quan TM, Nguyen-Duc T, Jeong W-K. Compressed Sensing MRI Reconstruction using a Generative Adversarial Network with a Cyclic Loss. IEEE Trans. Med. Imaging [Internet] 2017;37:1488–1497. doi: 10.1109/TMI.2018.2820120.

15. Mardani M, Gong E, Cheng JY, Vasanawala SS, Zaharchuk G, Xing L, Pauly JM. Deep Generative Adversarial Neural Networks for Compressive Sensing (GANCS) MRI. IEEE Trans. Med. Imaging [Internet] 2018:1–1. doi: 10.1109/TMI.2018.2858752.

16. Wang S, Su Z, Ying L, Peng X, Zhu S, Liang F, Feng D, Liang D, Technologies I. ACCELERATING MAGNETIC RESONANCE IMAGING VIA DEEP LEARNING. In: 2016 IEEE 13th International Symposium on Biomedical Imaging (ISBI). IEEE; 2016. pp. 514–517. doi: 10.1109/ISBI.2016.7493320.

17. Schlemper J, Caballero J, Hajnal J V., Price A, Rueckert D. A Deep Cascade of Convolutional Neural Networks for Dynamic MR Image Reconstruction. IEEE Trans. Med. Imaging [Internet] 2017:1–1. doi: 10.1007/978-3-319-59050-9\_51.

18. Han Y, Yoo J, Kim HH, Shin HJ, Sung K, Ye JC. Deep learning with domain adaptation for accelerated projection-reconstruction MR. Magn. Reson. Med. [Internet] 2018;80:1189–1205. doi: 10.1002/mrm.27106.

19. Eo T, Jun Y, Kim T, Jang J, Lee HJ, Hwang D. KIKI-net: Cross-domain convolutional neural networks for reconstructing undersampled magnetic resonance images. Magn. Reson. Med. [Internet] 2018. doi: 10.1002/mrm.27201.

20. Zhu B, Liu JZ, Cauley SF, Rosen BR, Rosen MS. Image reconstruction by domain-transform manifold learning. Nature [Internet] 2018;555:487–492. doi: 10.1038/nature25988.

21. Akçakaya M, Moeller S, Weingärtner S, Uğurbil K. Scan-specific robust artificial-neural-networks for \textit{k}-space interpolation (RAKI) reconstruction: Database-free deep learning for fast imaging. Magn. Reson. Med. [Internet] 2019;81:439–453. doi: 10.1002/mrm.27420.

22. Sodickson DK, Manning WJ. Simultaneous acquisition of spatial harmonics (SMASH): fast imaging with radiofrequency coil arrays. Magn. Reson. Med. [Internet] 1997;38:591–603.

23. Griswold MA, Jakob PM, Heidemann RM, Nittka M, Jellus V, Wang J, Kiefer B, Haase A. Generalized Autocalibrating Partially Parallel Acquisitions (GRAPPA). Magn. Reson. Med. 2002;47:1202–1210. doi: 10.1002/mrm.10171.

24. Pruessmann KP, Weiger M, Scheidegger MB, Boesiger P. SENSE: Sensitivity Encoding for Fast MRI. Magn Reson Med [Internet] 1999;42:952–962.

25. Lustig M, Donoho D, Pauly JM. Sparse MRI: The application of compressed sensing for rapid MR imaging. Magn Reson Med [Internet] 2007;58:1182–1195. doi: 10.1002/mrm.21391.

26. Lustig M, Pauly JM. SPIRiT: Iterative self-consistent parallel imaging reconstruction from arbitrary \textit{k}-space. Magn. Reson. Med. [Internet] 2010;64:457–471. doi: 10.1002/mrm.22428.

27. Otazo R, Kim D, Axel L, Sodickson DK. Combination of compressed sensing and parallel imaging for highly accelerated first-pass cardiac perfusion MRI. Magn. Reson. Med. [Internet] 2010;64:767–776. doi: 10.1002/mrm.22463.

28. Knoll F, Hammernik K, Kobler E, Pock T, Recht MP, Sodickson DK. Assessment of the generalization of learned image reconstruction and the potential for transfer learning. Magn. Reson. Med. [Internet] 2018. doi: 10.1002/mrm.27355.

29. Yang G, Yu S, Dong H, et al. DAGAN: Deep De-Aliasing Generative Adversarial Networks for Fast Compressed Sensing MRI Reconstruction. IEEE Trans. Med. Imaging [Internet] 2018;37:1310–1321. doi: 10.1109/TMI.2017.2785879.

30. Goodfellow IJI, Pouget-Abadie J, Mirza M, Xu B, Warde-Farley D, Ozair S, Courville A, Bengio Y. Generative Adversarial Networks. arXiv Prepr. arXiv … [Internet] 2014:1–9. doi: 10.1001/jamainternmed.2016.8245.

31. Zhu J-Y, Park T, Isola P, Efros AA. Unpaired Image-to-Image Translation Using Cycle-Consistent Adversarial Networks. In: 2017 IEEE International Conference on Computer Vision (ICCV). Vol. 2017–Octob. IEEE; 2017. pp. 2242–2251. doi: 10.1109/ICCV.2017.244.

32. Isola P, Zhu J-Y, Zhou T, Efros AA. Image-to-Image Translation with Conditional Adversarial Networks. ArXiv e-prints [Internet] 2016.

33. Feng L, Grimm R, Block KT, Chandarana H, Kim S, Xu J, Axel L, Sodickson DK, Otazo R. Golden-angle radial sparse parallel MRI: Combination of compressed sensing, parallel imaging, and golden-angle radial sampling for fast and flexible dynamic volumetric MRI. Magn. Reson. Med. [Internet] 2014;72:707–717. doi: 10.1002/mrm.24980.

34. Ronneberger O, Fischer P, Brox T. U-Net: Convolutional Networks for Biomedical Image Segmentation. In: Navab N, Hornegger J, Wells WM, Frangi AF, editors. Medical Image Computing and Computer-Assisted Intervention -- MICCAI 2015: 18th International Conference, Munich, Germany, October 5-9, 2015, Proceedings, Part III. Cham: Springer International Publishing; 2015. pp. 234–241. doi: 10.1007/978-3-319-24574-4\_28.

35. Gong E, Pauly JM, Wintermark M, Zaharchuk G. Deep learning enables reduced gadolinium dose for contrast-enhanced brain MRI. J. Magn. Reson. Imaging [Internet] 2018. doi: 10.1002/jmri.25970.

36. Han YS, Yoo J, Ye JC. Deep Residual Learning for Compressed Sensing CT Reconstruction via Persistent Homology Analysis. arXiv [Internet] 2016.

37. Ledig C, Theis L, Huszar F, et al. Photo-Realistic Single Image Super-Resolution Using a Generative Adversarial Network. ArXiv e-prints [Internet] 2016.

38. Li C, Wand M. Precomputed Real-Time Texture Synthesis with Markovian Generative Adversarial Networks. ArXiv e-prints [Internet] 2016.

39. He K, Zhang X, Ren S, Sun J. Delving Deep into Rectifiers: Surpassing Human-Level Performance on ImageNet Classification. ArXiv e-prints [Internet] 2015;1502.

40. Kingma DP, Ba J. Adam: A Method for Stochastic Optimization. ArXiv e-prints [Internet] 2014.

41. François Chollet. Keras. GitHub 2015:https://github.com/fchollet/keras.

42. Abadi M, Agarwal A, Barham P, et al. TensorFlow: Large-Scale Machine Learning on Heterogeneous Distributed Systems. ArXiv e-prints [Internet] 2016. doi: 10.1109/TIP.2003.819861.

43. Walsh DO, Gmitro AF, Marcellin MW. Adaptive reconstruction of phased array MR imagery. Magn. Reson. Med. [Internet] 2000;43:682–690. doi: 10.1002/(SICI)1522-2594(200005)43:5<682::AID-MRM10>3.0.CO;2-G.

44. Seiberlich N, Breuer F, Blaimer M, Jakob P, Griswold M. Self-calibrating GRAPPA operator gridding for radial and spiral trajectories. Magn. Reson. Med. [Internet] 2008;59:930–935. doi: 10.1002/mrm.21565.

45. Benkert T, Tian Y, Huang C, DiBella EVR, Chandarana H, Feng L. Optimization and validation of accelerated golden-angle radial sparse MRI reconstruction with self-calibrating GRAPPA operator gridding. Magn. Reson. Med. [Internet] 2018;80:286–293. doi: 10.1002/mrm.27030.

46. Bo Liu, Yi Ming Zou, Leslie Ying. Sparsesense: Application of compressed sensing in parallel MRI. In: 2008 International Conference on Technology and Applications in Biomedicine. IEEE; 2008. pp. 127–130. doi: 10.1109/ITAB.2008.4570588.

47. Krotkov EP. Active Computer Vision by Cooperative Focus and Stereo. New York, NY: Springer New York; 1989. doi: 10.1007/978-1-4613-9663-5.

48. Feng L, Huang C, Shanbhogue K, Sodickson DK, Chandarana H, Otazo R. RACER-GRASP: Respiratory-weighted, aortic contrast enhancement-guided and coil-unstreaking golden-angle radial sparse MRI. Magn. Reson. Med. [Internet] 2018;80:77–89. doi: 10.1002/mrm.27002.

49. Ma D, Gulani V, Seiberlich N, Liu K, Sunshine JL, Duerk JL, Griswold MA. Magnetic resonance fingerprinting. Nature [Internet] 2013;495:187–192. doi: 10.1038/nature11971.

50. Ye JC, Han Y, Cha E. Deep Convolutional Framelets: A General Deep Learning Framework for Inverse Problems. ArXiv e-prints [Internet] 2017.

51. Mao X, Li Q, Xie H, Lau RYK, Wang Z, Smolley SP. Least Squares Generative Adversarial Networks. ArXiv e-prints [Internet] 2016.

52. Gulrajani I, Ahmed F, Arjovsky M, Dumoulin V, Courville A. Improved Training of Wasserstein GANs. ArXiv e-prints [Internet] 2017.

53. Mirza M, Osindero S. Conditional Generative Adversarial Nets. ArXiv e-prints [Internet] 2014.

54. Karras T, Aila T, Laine S, Lehtinen J. Progressive Growing of GANs for Improved Quality, Stability, and Variation. ArXiv e-prints [Internet] 2017.

\end{document}